\documentclass[10pt,twocolumn,letterpaper]{article}

\usepackage{iccv}
\usepackage{times}
\usepackage{epsfig}
\usepackage{graphicx}
\usepackage{amsmath}
\usepackage{amssymb}

\usepackage{multirow}
\usepackage{array}
\usepackage{makecell}
\usepackage{stfloats}
\usepackage{caption}
\usepackage{subcaption}

\usepackage[pagebackref=true,breaklinks=true,letterpaper=true,colorlinks,bookmarks=false]{hyperref}

\iccvfinalcopy 

\def\iccvPaperID{4} 

\ificcvfinal\fi

\begin{document}

\title{
Unmanned Aerial Vehicle Visual Detection and Tracking using Deep Neural Networks: A Performance Benchmark
}


\author{\hspace{-1.5cm}Brian K. S. Isaac-Medina\\
 \hspace{-1.5cm}Department of Computer Science\\
 \hspace{-1.5cm}Durham University\\
\hspace{-1.5cm}Durham, UK\\
\hspace{-1.5cm}{\tt\small brian.k.isaac-medina@durham.ac.uk}

\and
\hspace{1.5cm}Matt Poyser \\
 \hspace{1.5cm}Department of Computer Science\\
 \hspace{1.5cm}Durham University\\
\hspace{1.5cm}Durham, UK\\
\hspace{1.5cm}{\tt\small matthew.poyser@durham.ac.uk}

\and
\hspace{-2cm}Daniel Organisciak\\
 \hspace{-2cm}Department of Computer and Information Sciences\\
 \hspace{-2cm}Northumbria University\\
\hspace{-2cm}Newcastle upon Tyne, UK\\
\hspace{-2cm}{\tt\small daniel.organisciak@northumbria.ac.uk}

\and 
Chris G. Willcocks\\
 Department of Computer Science\\
 Durham University\\
Durham, UK\\
{\tt\small christopher.g.willcocks@durham.ac.uk}

\and 
\hspace{-1.5cm}Toby P. Breckon\\
\hspace{-1.5cm} Department of Computer Science\\
 \hspace{-1.5cm}Durham University\\
\hspace{-1.5cm}Durham, UK\\
\hspace{-1.5cm}{\tt\small toby.breckon@durham.ac.uk}

\and 
\hspace{2cm}Hubert P. H. Shum\\
 \hspace{2cm}Department of Computer Science\\
 \hspace{2cm}Durham University\\
\hspace{2cm}Durham, UK\\
\hspace{2cm}{\tt\small hubert.shum@durham.ac.uk}

}





\maketitle
\ificcvfinal\thispagestyle{empty}\fi

\begin{abstract}

Unmanned Aerial Vehicles (UAV) can pose a major risk for aviation safety, due to both negligent and malicious use. For this reason, the automated detection and tracking of UAV is a fundamental task in aerial security systems. Common technologies for UAV detection include visible-band and thermal infrared imaging, radio frequency and radar. Recent advances in deep neural networks (DNNs) for image-based object detection open the possibility to use visual information for this detection and tracking task. Furthermore, these detection architectures can be implemented as backbones for visual tracking systems, thereby enabling persistent tracking of UAV incursions. To date, no comprehensive performance benchmark exists that applies DNNs to visible-band imagery for UAV detection and tracking. To this end, three datasets with varied environmental conditions for UAV detection and tracking, comprising a total of 241 videos (331,486 images), are assessed using four detection architectures and three tracking frameworks. The best performing detector architecture obtains an mAP of 98.6\% and the best performing tracking framework obtains a MOTA of 98.7\%. Cross-modality evaluation is carried out between visible and infrared spectrums, achieving a maximal 82.8\% mAP on visible images when training in the infrared modality. These results provide the first public multi-approach benchmark for state-of-the-art deep learning-based methods and give insight into which detection and tracking architectures are effective in the UAV domain.

\end{abstract}


\section{Introduction}

Over the past decade, Unmanned Aerial Vehicles  (UAV) have become more accessible for both professional and casual use, resulting in an increased risk of UAV-caused disturbances. Whether a pilot has malicious intent or is negligent, UAVs can impact aviation safety, travel into restricted airspace, or capture sensitive data. UAV disturbances can also have serious security and economic consequences \cite{cuas_2019}. These disturbances are more likely to occur as UAV usage grows.

As such, a major focus of counter-UAV systems is to detect and track UAVs that are small and fast-moving. Traditional technologies used in counter-UAV systems include radio frequency \cite{radio_freq_xiao}, radar \cite{radar_klare} and acoustic sensors \cite{acoustics_yang}. These sensors, however, perform sub-optimally to locate objects at long distances \cite{ml_drone_detection_review, short_distance_radar, jeon2017empirical} or in settings with significant sources of noise such as airports, where UAV threats are prevalent \cite{defending_airports, scheller_2017, japcc_2019}. While the use of standard cameras is attractive due to the visual cues they provide to security personnel, a key challenge is the relatively small image size of UAVs, which impacts the detection performance. With the recent advances in deep neural networks, image-based models have become widely used for video surveillance \cite{surveillance}. Furthermore, these models can be used in infrared (IR) imagery, overcoming common problems when using optical images, such as lighting or weather conditions, although the data is more limited. To this end, we focus our research on vision-based systems.

In the last decade, deep convolutional neural networks (DCNN) have become the base model of state-of-the-art architectures for visual tasks such as object detection and tracking \cite{obj_det_dl, tracking_review}. These learning-based methods rely on extracting feature maps from the input images to provide a probability distribution over a set of categories or to regress a set of real-valued variables. In the context of object detection, DCNNs are used to regress a parametric model of bounding boxes enclosing the objects of interest \cite{fastrcnn, ssd, yolov3}, while more recent models use encoder-decoder architectures for the same task \cite{detr}. When operating within the tracking domain, objects must be assigned an additional identification feature. Subsequently, the tracking task involves further identification of the same instance at different locations in the temporal dimension. Most object tracking architectures are based on \emph{tracking-by-detection}, which consists of associating the predicted objects from standard detection architectures across the temporal dimension (i.e., consecutive video frames) \cite{Ciaparrone_2020}. 


The different settings in which counter-UAV systems may be encountered imply that highly robust tracking and detection models are required. 
According to Rozantsev \etal (2015) \cite{uav3}, there are three key challenges for such models. Firstly, given the speed of UAVs, they must be detected at a long distance, when they appear very small within a camera image. This problem is compounded by the fact that secondly, the background scene is often highly complex and diverse; landscapes can vary dramatically and also be highly cluttered. Thirdly, UAV are capable of relatively complex movement, and tracking them often involves cameras mounted on a high-speed system themselves.
Therefore, the evaluation of detection and tracking models over different datasets, demonstrating their robustness over such challenges, is an imperative task.

Although detecting and tracking UAV has been previously studied \cite{unlu2019deep, nalamati2019drone}, to the best of our knowledge, there has been no wide-ranging performance benchmark study evaluated across multiple datasets for these tasks. Concurrent to this work, Jiang \etal \cite{jiang2021antiuav} introduce the Anti-UAV dataset and benchmark tracking performance. However, they do not compare different detectors nor datasets, which can greatly impact tracking performance. As a result, it is difficult to evaluate the effectiveness of the proposed frameworks. 

Given this motivation, this work compares the performance of several vision-based detection and tracking models under three different datasets, subject to varying setups. Our main contributions are as follow:

\begin{itemize}
	\item[--] a benchmark performance study is presented across leading state of the art object detection (Faster-RCNN \cite{fasterrcnn}, YOLOv3 \cite{yolov3}, SSD \cite{ssd}, and DETR \cite{detr}) and tracking architectures (SORT \cite{sort}, DeepSORT \cite{deepsort}, and Tracktor \cite{tracktor}).
	\item[--] state-of-the-art detection performance (0.986 mAP) is achieved for visual detection and tracking within the counter-UAV domain, compared to previous leading performance (0.952 mAP \cite{yolo_uav_detection}) .
	\item[--] collation and assessment of three UAV datasets (MAV-VID \cite{mav-vid}, Drone-vs-Bird \cite{coluccia2019drone}, and Anti-UAV \cite{jiang2021antiuav}), yielding a plethora of varied environment and environmental condition detection challenges. Images are captured by both ground and UAV-mounted cameras, under highly dynamic scenes, at short and long distances and by both optical and IR cameras.
	\item[--] novel evaluation of cross-modality training and testing for UAV. Training on IR data is demonstrated to be sufficient for inference on RGB imagery (0.828 mAP).
\end{itemize}

Our benchmark toolkit is open for further research and development. It can be downloaded on the project website: \url{https://github.com/KostadinovShalon/UAVDetectionTrackingBenchmark}

\section{UAV Detection and Tracking Architectures}
Overall, prior work within the UAV domain is limited in both scope and complexity whereas multiple comprehensive surveys for detection and tracking have already been published more broadly \cite{obj_det_dl} \cite{tracking_review}. We provide an overview of popular architectures and literature that addresses generic detection and tracking challenges. 

\subsection{Object Detection} 
Object detection is one of the most fundamental and well-studied tasks in computer vision. Objects from desired classes must be located with a bounding box, which should enclose the object as tightly as possible. Deep learning has strongly influenced object detectors because of the variety of expressive features that deep models can learn, leading towards a focus on generic object detectors that perform well on a number of distinct classes and datasets, rather than designing detectors for specific objects \cite{liu2020deep}. 

Deep-learning architectures are mostly composed of two categories: two-stage detectors and one-stage detectors. The two-stage detectors, such as RCNN \cite{rcnn}, Faster RCNN \cite{fasterrcnn} and RFCN \cite{rfcn}, consist of one stage to compose a set of candidate regions of interest that could contain desired objects and a second stage to classify the proposed regions and regress its bounding box parameters. These detectors typically outperform one-stage detection architectures in terms of detection accuracy at the expense of computational efficiency.

One-stage detectors such as OverFeat \cite{overfeat}, YOLO (You Only Look Once) \cite{yolov3}, SSD (Single Shot Detector) \cite{ssd} and CornerNet \cite{cornernet} drop the region proposal stage, favouring the use of global image features to determine bounding box locations. In particular, YOLO and SSD split the image into a grid and regress the parameters of the bounding box with respect to a set of anchor boxes on each grid cell. CornerNet instead focuses on detecting pairs of keypoints at the top-left and bottom-right of each object. DETR (Detection Transformer) \cite{detr} is a more recent approach that uses a transformer encoder-decoder architecture with a bipartite matching loss to alternatively propose bounding boxes.

Despite the considerable progress on object detection, a major challenge is that UAV often appear at dramatically different sizes in the image. 
Small objects are composed of fewer pixels and subsequently the deepest layers of the convolutional backbone within such detection architectures can struggle to extract detailed object information. Furthermore, successive max-pooling layers may suppress such detection responses for small scene objects within the penultimate stages of the architecture. To combat this, feature pyramid networks \cite{featurepyramids} fuse features from several layers to perform detection at multiple scales. 
An additional problem is differentiating UAV from other small, similarly-coloured objects that appear in the sky, such as birds. This is epitomised by the Drones vs Birds dataset \cite{coluccia2019drone}, where the winner team implemented a two-stage detector having temporal-aware input channels and a standard tracking algorithm to filter out false positives \cite{craye2019spatio}.

UAV can also be difficult to detect due to poor visibility in adverse weather conditions, poor lighting, low-quality cameras (on other UAV) and buildings with a similar colour profile. 
Yang \etal \cite{yang2020advancing} survey methods to improve visibility for detection, including de-hazing, de-raining and low-light enhancement. These methods improve detection results on well-researched objects such as faces, pedestrians and vehicles, but it is unclear whether they transfer to UAV.

\subsection{Tracking}
Single-object tracking (SOT) introduces the key challenge of distinguishing objects from the background, for which relatively established traditional algorithms can perform very well \cite{condensation, condensationplus, ensemble, siameseSOT}. Multi-object tracking (MOT), however, within which lies the scope of this paper, introduces further challenges. In particular, we must be able to track multiple objects that occlude each other over a short or long time frame. To this end, algorithms commonly employ an appearance descriptor \cite{appearance1, appearance2, appearance3} or exploit motion information \cite{motion1, motion2} to differentiate between nearby moving objects. 

Geiger \etal \cite{geiger} employ the Hungarian Algorithm \cite{hungarian} for object tracking. They are able to precisely track vehicles within urban scenes by associating predicted locations from motion, with detected locations and object geometry from appearance cues. Bewley \etal \cite{sort} demonstrate that good performance can be achieved (but not limited to vehicle tracking) without the need of expensive geometric calculations. Additionally, they demonstrate that their online framework is quicker than other methods, without sacrificing accuracy. Wojke \etal (DeepSORT) \cite{deepsort} further introduce a deep auxiliary network within the SORT framework to better capture an appearance descriptor of the objects. This re-identification network, robust to object pose and camera viewpoint changes, is used within the association step to dramatically reduce the number of times an object is identified as new rather than associated with one that has been previously identified. Bergmann \etal (Tracktor) \cite{tracktor} indicate that detectors are in fact enough for a tracking system, and that there is no need for explicit motion prediction and association. They show that motion models can be sufficiently captured within the bounding box regression of detectors, and achieve state of the art accuracy and frame throughput.

In almost all cases however, the detection model architecture exists as a backbone within the tracking framework. Therefore, evaluating tracking performance is a natural step in determining the general performance of a detector, which we include in this benchmark.


\section{Experimental Setup}

\subsection{UAV Datasets} \label{ssec:datasets}

\begin{figure*}[t]
\centering
\includegraphics[width=0.8\linewidth]{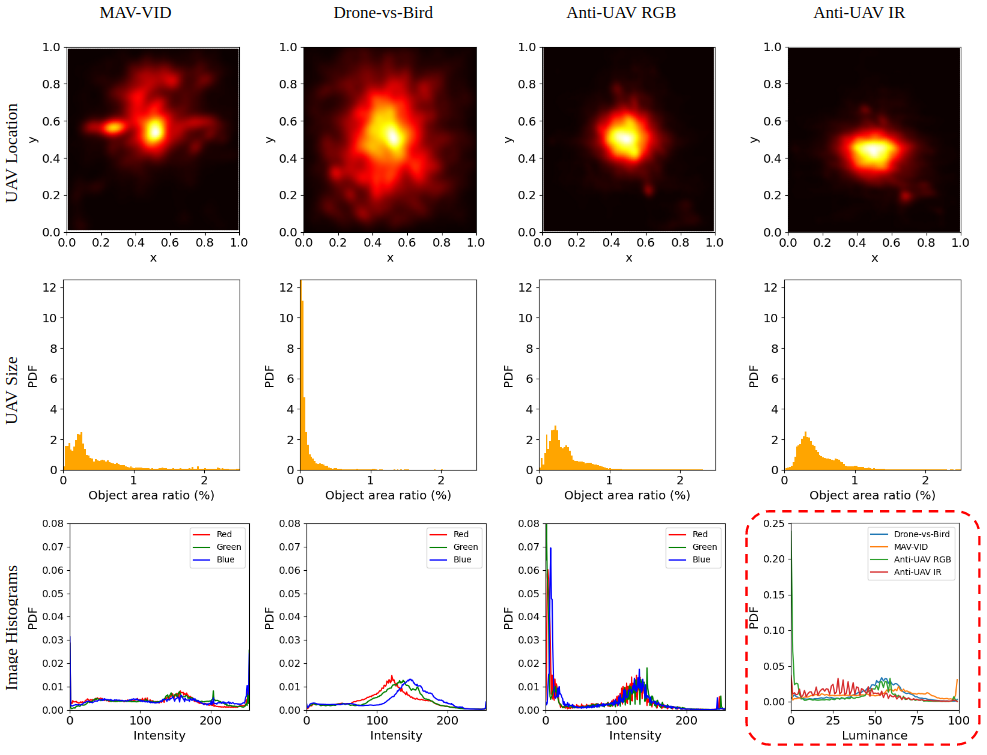}
\caption{Dataset statistics. UAV location refers to the bounding box centre of the objects in the image. UAV size is the ratio of the object size with the image size. The colour histograms are generated using all images from the dataset. The image in the red dashed square shows the luminance component from the L*a*b* colour space of each dataset.}
\label{Fig:dataset_statistics}
\end{figure*}

In order to evaluate the performance of DCNNs based architectures for object detection and consequently tracking, three datasets are assessed in this work. Each dataset comprises a set of varying length videos subject to different constraints and setups. These datasets have only one class, labelled as ``drone''. Figure \ref{Fig:dataset_statistics} shows the datasets statistics.

\textbf{Multirotor Aerial Vehicle VID (MAV-VID) \cite{mav-vid}}. This dataset consists on videos at different setups of single UAV. It contains videos captured from other drones, ground based surveillance cameras and handheld mobile devices. It comprises 53 videos (29,500 images) for training and 11 videos (10,732 images) for validation, with an average object size of $136\times77$ pixels (0.66\% of the image size). Although this dataset can be considered as the most straightforward among the three, the differences between the videos make this dataset a good benchmark to evaluate the ability of detectors and trackers for generalisation. As seen in Figure \ref{Fig:dataset_statistics}, UAVs usually move across the $x$ axis and are recorded from the bottom.

\textbf{Drone-vs-Bird Detection Challenge \cite{coluccia2019drone}}. As part of the International Workshop on Small-Drone Surveillance, Detection and Counteraction techniques of IEEE AVSS 2020, the main goal of this challenge is to reduce the high false positive rates that vision-based methods usually suffer. This dataset comprises videos of UAV captured at long distances and often surrounded by small objects, such as birds, with an average UAV in image size of $34\times23$ pixels (0.10\% of the image size). This dataset is formed by 77 videos (104,760 images), which we divided in 61 videos (85,904 images) for training and 16 videos (18,856 images) for validation. Figure \ref{Fig:dataset_statistics} reveals that UAVs in this dataset are smaller and appear with more variation across the image plane.

\textbf{Anti-UAV \cite{jiang2021antiuav}}. This multi-modal dataset comprises 100 fully-annotated RGB and IR unaligned videos, giving a total of 186,494 images including both modalities. Anti-UAV dataset is intended to provide a real-case benchmark for evaluating object tracker algorithms in the context of UAV. It contains recordings of 6 UAV models flying at different lightning and background conditions. The average object size for the RGB modality is $125\times59$ pixels (0.40\% of the image size) and $52\times29$ pixels (0.50\% of the image size). We divided this dataset in 60 videos (149,478 images) for training and 40 videos (37,016) for validation. In our experiments, we refer to \emph{Anti-UAV Full} to the dataset containing all images regarding their modality while \emph{Anti-UAV RGB} and \emph{Anti-UAV IR} refer to each separate modality. It is apparent from Figure \ref{Fig:dataset_statistics} that UAVs are slightly bigger relative to the image size. Similar intensities of the three colour channels indicate the poor contrast in the RGB videos. Furthermore, the shape of the histogram indicates a possible pre-processing carried out during the acquisition process.

\begin{figure*}[t]
    \centering
    \begin{subfigure}[b]{0.48\textwidth}
    	\includegraphics[width=\linewidth]{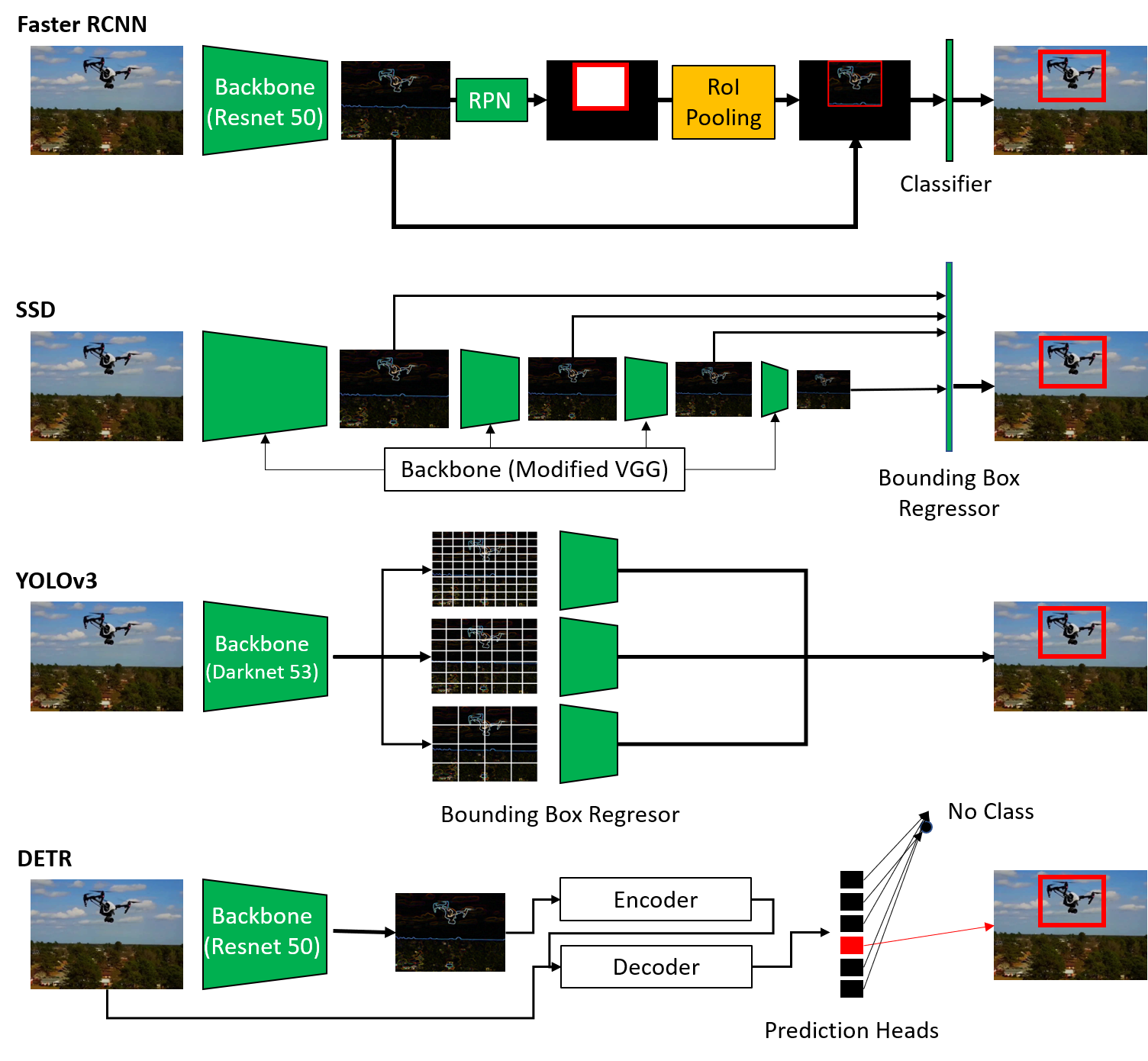}
    	\centering
    	\label{Fig:detectors}
    \end{subfigure}
    \hspace{1em}%
    \begin{subfigure}[b]{0.48\textwidth}
        \includegraphics[width=\linewidth]{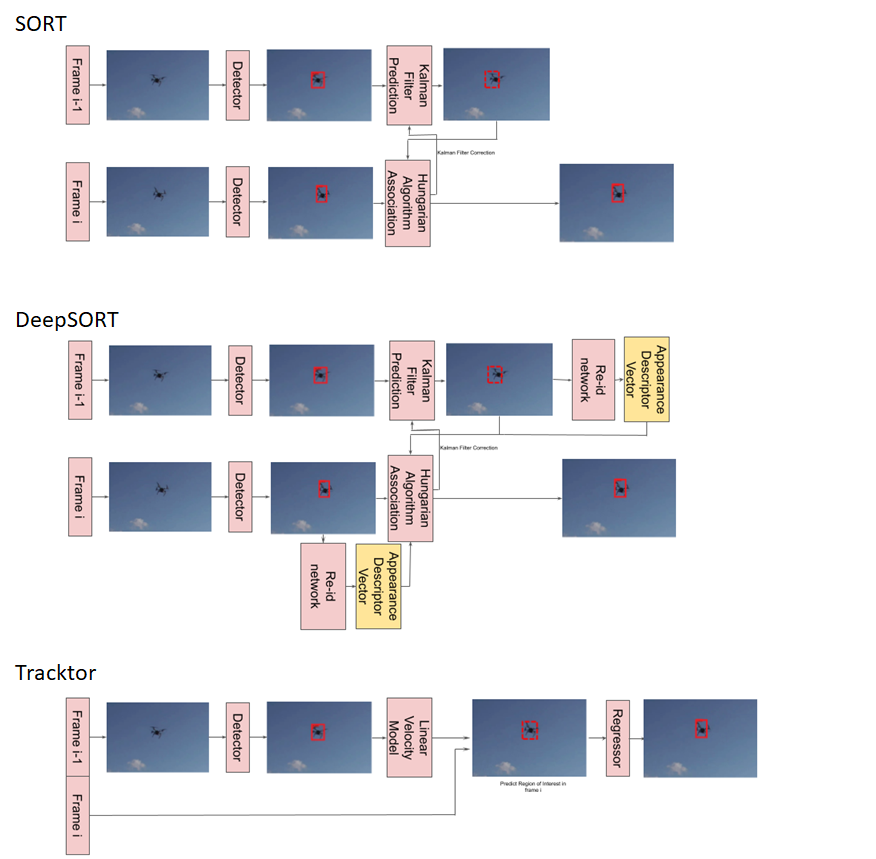}
        \centering
        \label{Fig:trackers}
    \end{subfigure}
    \caption{Architectures evaluated in this work. Detectors (left): Faster RCNN, SSD, YOLOv3 and DETR. Tracking frameworks (right): SORT, DeppSORT and Tracktor.}
    \label{Fig:architectures}
\end{figure*}

The plot with the red dashed square in Figure \ref{Fig:dataset_statistics} shows the luminance (\emph{L} from the L*a*b* colour space) of each dataset. Anti-UAV IR has the lowest luminance across the datasets, which is expected from an IR dataset. On the other hand, MAV-VID shows a higher luminance, indicating it is brighter than the other datasets.

\subsection{Object Detection} \label{ssec:objdet}
With the aim of assessing the detection performance, we evaluate four detectors corresponding to different object detection paradigms: Faster RCNN \cite{fasterrcnn}, SSD512 \cite{ssd}, YOLOv3 and DETR \cite{detr}. All networks are pretrained on the MS COCO dataset \cite{coco_dataset}, a general-purpose dataset containing more than 330k images and 1.5M instances over 80 categories.  Regarding the optimisation algorithm, all models were trained for 24 epochs, decreasing the learning rate by a factor of 0.1 after epochs 16 and 22. A brief diagram for each detector architecture used in this work is shown in Figure \ref{Fig:architectures}.

\textbf{Faster RCNN} \cite{fasterrcnn} is trained using a ResNet-50 \cite{resnet} as backbone and a Feature Pyramid Network \cite{featurepyramids} at the end of each convolutional block (conv2\_3, conv3\_4, conv4\_6 and conv5\_3) with 256 output channels on each level. Batch normalization is used for regularisation and stochastic gradient descent with a learning rate of 0.001. 

\textbf{SSD512} \cite{ssd} is implemented with the same configuration details in the original work and with the same optimisation settings as in Faster RCNN. 

\textbf{YOLOv3} \cite{yolov3}, from the YOLO family of detectors, is trained using a DarkNet-53 backbone. Input images were resized and square padded to have a final size of $608\times608$. The same optimisation technique as in Faster RCNN and SSD512 was used. 

\textbf{DETR} \cite{detr}, one of the most recent architectures for detection, is trained with a ResNet-50 backbone and the original implementation details (AdamW \cite{adamw} as the optimisation algorithm with an initial learning rate of $10^{-4}$). However, the learning rate was decreased by a factor of 10 for the Drone-vs-Bird dataset to achieve convergence.

Detection performance is evaluated using MS COCO metrics. In order to compare our results with other works, we will refer the COCO AP$_{0.5}$ simply as \emph{mAP}. All models were trained using the MMDetection framework \cite{mmdetection} with an Nvidia 2080 Ti GPU.

\subsection{Object Tracking}
To evaluate object tracking within the counter-UAV domain, we employ each of the detectors detailed in section \ref{ssec:objdet}, within three MOT frameworks: SORT \cite{sort}, DeepSORT \cite{deepsort}, and Tracktor \cite{tracktor}. While several of the datasets consist of only a single UAV within each frame, and thus SOT trackers would suffice, we consider only MOT trackers that are more generalizable to counter-UAV domains.

\textbf{SORT} was implemented using the same configuration as the original paper from Bewley \etal \cite{sort}. As such, a Kalman filter is employed to capture the linear velocity motion model of the UAV, with the Hungarian algorithm to associate detected and predicted tracks.

Our \textbf{DeepSORT} framework builds upon SORT as per the original work by Wojke \etal \cite{deepsort}. Thus we incorporate a pre-trained deep re-identification network (that can be found within the MMTracking toolbox) to generate appearance descriptors. This network has been trained by the Tracktor authors and is equivalently usable within the DeepSORT framework. Additional DeepSORT tracking details can be found within the original paper. Otherwise, implementation details remain the same as SORT.

\textbf{Tracktor} utilizes a camera motion compensation model to generate track predictions, alongside the same re-identification network employed within DeepSORT. Other implementation details are unchanged from the original paper from Bergmann \etal \cite{tracktor}.
 
We test the performance of each tracking framework under the UAV datasets illustrated in \ref{Fig:architectures}, with respect to multi object tracking accuracy (MOTA \cite{mota}), the ratio of correctly identified detections over the average number of ground-truth and computed detections (IDF1 \cite{idf1}), and how often a detected object is assigned an incorrect ID (`ID Sw.'). Each tracker was implemented using the MMTracking toolbox\footnote{Marginal changes are required to the MMTracking toolbox to enable single stage detectors within Tracktor (as per 05/03/2021)}. 



\section{Results}
In this section we review the results for both detection and tracking tasks. These results correspond to the best performing detectors when evaluated over the validation sets; the same maximally performing weights are used within the tracking evaluation.

\subsection{Detection} \label{ssec:results-detection}
Comparative UAV detection performance among the datasets and detectors is shown in Table \ref{Table:detection_results}. It can be observed that detection performance varies significantly across the datasets with the best performance achieved in the Anti-UAV RGB dataset (0.986 mAP). This dataset has the most uniform set of image sequences since all videos are recorded in similar settings, with only 6 types of UAV present within it. The mAP across both modalities in the Anti-UAV dataset remains similar (mAP ranges from 0.978 to 0.986 for RGB and from 0.975 to 0.980 for IR), implying that the detectors are learning to capture the shape rather than the colour information. This is further supported by the cross-modality experiments, discussed later. However, the IR dataset obtains a dramatic increase of the AP for small objects (maximal  AP$_S$: 0.533 via YOLOv3). This can be explained by the higher contrast for images captured with harsh weather conditions or poor lighting. Detection performance for the MAV-VID dataset is similar to Anti-UAV (0.978 mAP with Faster RCNN). Although MAV-VID dataset comprises bigger and clearer UAV images, it has less input data than the other two datasets. Additionally, the low precision for small objects can be explained by the low number of visually small UAV in the training set. This reveals the need of annotated datasets with UAV at a long distance. Finally, the Drone-vs-Bird dataset obtains a maximal mAP of 0.667 via DETR and COCO AP of 0.283 via Faster RCNN. This indicates that Faster RCNN is better at predicting tighter bounding boxes. The relatively poorer performance in the Drone-vs-Bird dataset can be attributed to the significantly smaller size of the UAVs (Figure \ref{Fig:dataset_statistics}).


\begin{table*}[!t]
    \centering
    \caption{Detection Performance Benchmark}
    \label{Table:detection_results}
    \resizebox{0.8\linewidth}{!}{%
    \begin{tabular}{c | c | c c c c c c| c c c c}
    \Xhline{2\arrayrulewidth}
    \textbf{Dataset} & \textbf{Model} & $\bf{AP}$ & $\bf{AP_{0.5}}$& $\bf{AP_{0.75}}$ & $\bf{AP_S}$ & $\bf{AP_{M}}$ & $\bf{AP_{L}}$ 
    & $\bf{AR}$ & $\bf{AR_{S}}$ & $\bf{AR_M}$ & $\bf{AR_L}$ \\
    \Xhline{2\arrayrulewidth}
    \multirow{4}{*}{MAV-VID} & Faster RCNN & \bf 0.592  & \bf 0.978  & \bf 0.672  & \bf 0.154 & \bf 0.541  & \bf 0.656  & 0.659 & 0.369  & 0.621  & 0.721 \\
    
    & SSD512 & 0.535  & 0.967  & 0.536  & 0.083 & 0.499  &  0.587  & 0.612 & \bf 0.377  & 0.578  & 0.666 \\
    
    & YOLOv3 & 0.537  & 0.963  & 0.542  & 0.066  & 0.471  & 0.636  & 0.612  &  0.208  &  0.559  & 0.696 \\
    
    & DETR & 0.545  & 0.971  & 0.560  & 0.044  & 0.490  & 0.612  & \bf 0.692  &  0.346  &  \bf 0.661  & \bf 0.742 \\
    \hline
    \multirow{4}{*}{Drone-vs-Bird} & Faster RCNN & \bf 0.283  & 0.632  & \bf 0.197  & \bf 0.218 & \bf 0.473  & 0.506 & 0.356  & 0.298  & 0.546  & 0.512 \\
    
    & SSD512 & 0.250  & 0.629  & 0.134  & 0.199 & 0.422  & 0.052  & 0.379 & 0.327  & 0.549  & 0.556 \\
    
    & YOLOv3 & 0.210  & 0.546  & 0.105  & 0.158  & 0.395  & 0.356  & 0.302  &\  0.238  &  0.512  & \bf 0.637 \\
    
    & DETR & 0.251  & \bf 0.667  & 0.123  & 0.190  & 0.444  & \bf 0.533  & \bf 0.473  & \bf 0.425  & \bf 0.631 & 0.550 \\
    \hline
    \multirow{4}{*}{Anti-UAV Full} & Faster RCNN  & 0.612  & 0.974  & \bf 0.701 & 0.517  & \bf 0.619  & \bf 0.737 & 0.666  & 0.601  & 0.670 & 0.778 \\
    
    & SSD512 & \bf 0.613  & \bf 0.982  & 0.697 & 0.527  & \bf 0.619  & 0.712  & \bf 0.678  & 0.616  & \bf 0.682  & \bf 0.780  \\
    
    & YOLOv3 & 0.604  & 0.977  & 0.676 & \bf 0.529  & \bf 0.619  & 0.708  & 0.667  & \bf 0.618  & 0.668  & 0.760  \\
    
    & DETR & 0.586  & 0.977 & 0.648  & 0.509  & 0.589  & 0.692  & 0.649  & 0.598  & 0.649 & 0.752  \\
    \hline
    \multirow{4}{*}{Anti-UAV RGB} & Faster RCNN & \bf 0.642  & 0.982  & \bf 0.770  & 0.134 & \bf 0.615  & 0.718  & 0.694  & 0.135  & 0.677  & 0.760  \\
    
    & SSD512 & 0.627  & 0.979  & 0.747 & 0.124  & 0.593  & 0.718  & \bf 0.703  & 0.156  & \bf 0.682  & 0.785  \\
    
    & YOLOv3 & 0.617  & \bf 0.986  & 0.717 & \bf 0.143  & 0.595  & 0.702  & 0.684 & \bf 0.181  & 0.664  & 0.758  \\
    
    & DETR & 0.628  & 0.978  & 0.740 & 0.129  & 0.590  & \bf 0.734  & 0.700  & 0.144  & 0.675  & \bf 0.794  \\
    \hline
    \multirow{4}{*}{Anti-UAV IR} & Faster RCNN & 0.581  & 0.977  & 0.641  & 0.523 & 0.623  & -  & 0.636  & 0.602  & 0.663  & -  \\
    
    & SSD512 & 0.590  & 0.975  & 0.639 & 0.518  & 0.636  & -  & 0.649  & 0.609  & 0.681  & -  \\
    
    & YOLOv3 & 0.591  & 0.976  & 0.643 & \bf 0.533  & 0.638  & -  & 0.651  & 0.620  & 0.675  & -  \\
    
    & DETR & \bf 0.599  & \bf 0.980  & \bf 0.655 & 0.525 & \bf 0.642  & -  & \bf 0.671  & \bf 0.633  & \bf 0.701  & -  \\
    \Xhline{2\arrayrulewidth}
    \end{tabular}%
    }
\end{table*}

\begin{table}[!ht]
    \centering
    \caption{Detection inference time}
    \label{Table:inference_time}
    \begin{tabular}{c | c | c | c}
    \Xhline{2\arrayrulewidth}
    \textbf{Model} & \textbf{GFLOPS} & \textbf{FPS} & \textbf{Params} \\ 
    \Xhline{2\arrayrulewidth}
    Faster RCNN & 207 & 18.0 & 41M \\
    SSD512 & 88 & 32.4 & 24M \\
    YOLOv3 & 70 & 36.0 & 61M \\
    DETR & 86 & 21.4 & 41M \\
    \Xhline{2\arrayrulewidth}
    \end{tabular}
\end{table}
Faster RCNN and DETR generally outperform the one-stage detectors. However, it is important to note that both Faster RCNN and DETR have a minimum image height of 800 pixels which allows for a better performance at detecting small UAV. On the other hand, the inference time for SSD512 and YOLOv3 models are faster with 32.4 and 36.0 fps, as seen in Table \ref{Table:inference_time}. Although YOLOv3 is the fastest architecture, SSD512 is the lightest model with only 24M parameters ($\sim$192 MB). As a result, in settings where memory and computational power is a constraint (such as an embedded system), lightweight networks such as SSD or YOLOv3 may be preferred.

Lastly, we evaluate the detector performance on RGB and IR modalities when testing on a different modality to the one that the network was trained on (Table \ref{Table:rgb_vs_ir_detection}). The results show that features learned in the IR domain can be used as a good approximation for detecting drones in the visual spectrum (0.828 mAP with Faster RCNN). This may indicate that the network is learning to represent a UAV based on its shape and not in its colour or texture. On the other hand, the Anti-UAV IR mAP dropped to 0.644 with Faster RCNN trained on the Anti-UAV RGB. This bigger may be caused by the loss of the colour information in IR imagery. One-stage detectors performed poorer, indicating the difficulty of getting the proper features on each grid cell defined by these architectures.

\subsection{Tracking}
We compare the performance of each detector within each tracking framework in Table \ref{Table:tracking_results_deepsort}. It can be observed that tracking performance varies dramatically, but largely depends upon the detection architecture employed. Evidently, Faster-RCNN once again clearly outperforms other detection models for the easier MAV-VID challenge (MOTA: 0.955), while SSD512 performs best overall for the Drone-vs-Bird (MOTA: 0.525) and Yolov3 for Anti-UAV (up to MOTA: 0.985) challenges. However, DETR still yields highest performance towards tracking small objects (better shown in the detection results in Table \ref{ssec:results-detection}). The Tracktor tracking framework overall yields best performance, owing to its vastly better ability to correctly attribute ids.

\begin{figure*}[t]
	\centering
	\begin{subfigure}[b]{0.22\textwidth}
		\includegraphics[width=\linewidth]{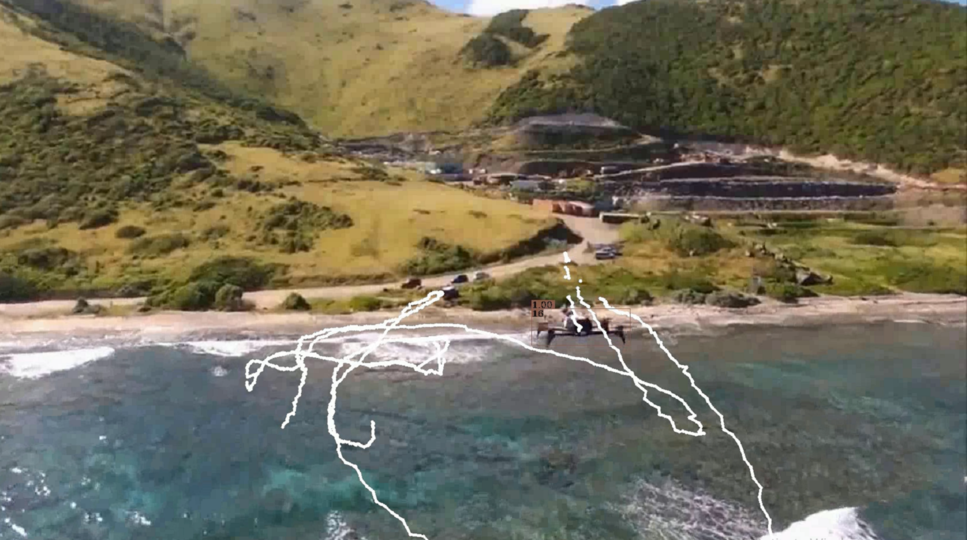}
		\centering
		\caption{MAV-VID}
		\label{Fig:mavvidtrack}
	\end{subfigure}
	\begin{subfigure}[b]{0.22\textwidth}
		\includegraphics[width=\linewidth]{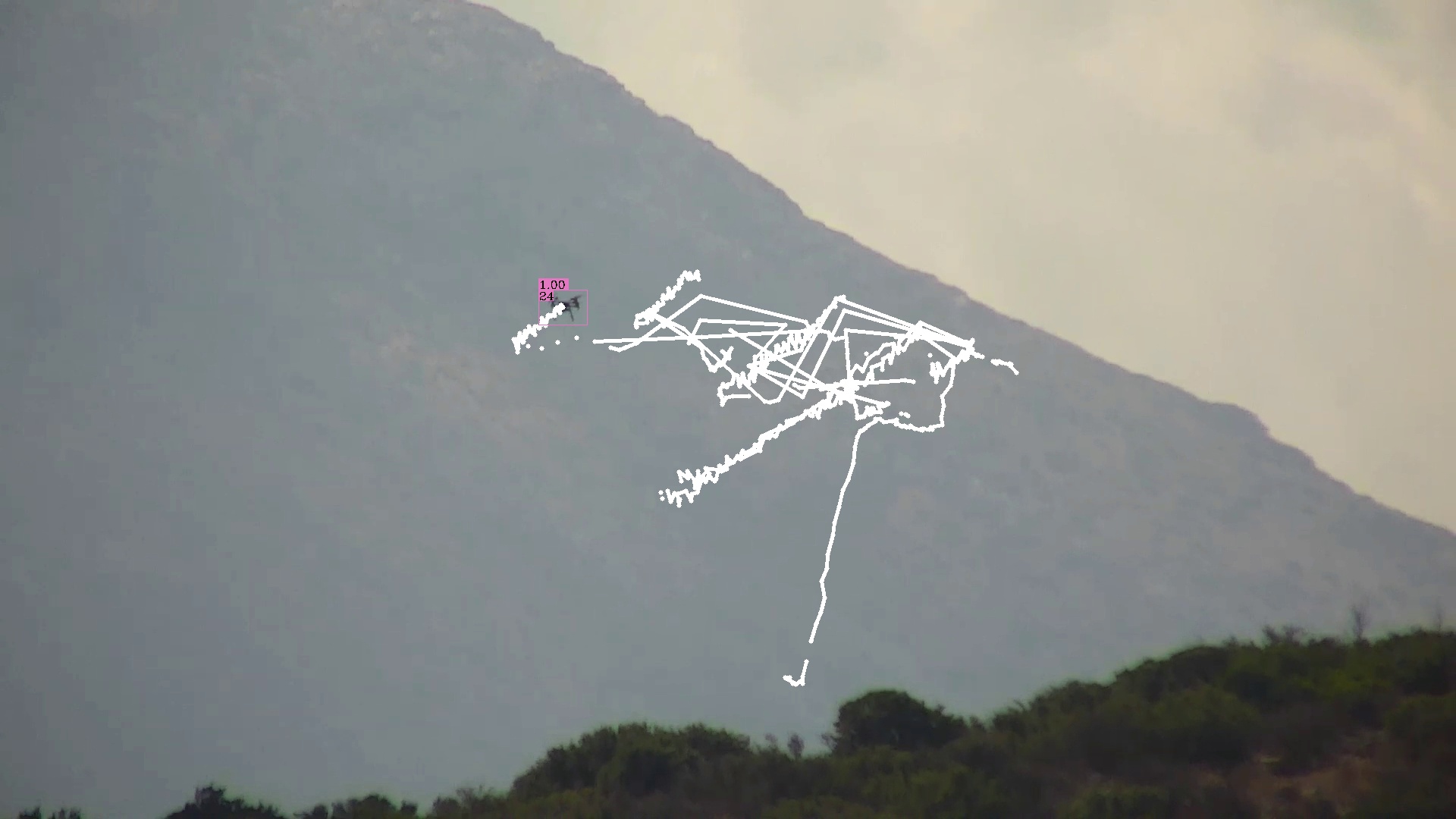}
		\centering
		\caption{Drone-vs-Bird}
		\label{Fig:birdtrack}
	\end{subfigure}
	\begin{subfigure}[b]{0.22\textwidth}
		\includegraphics[width=\linewidth]{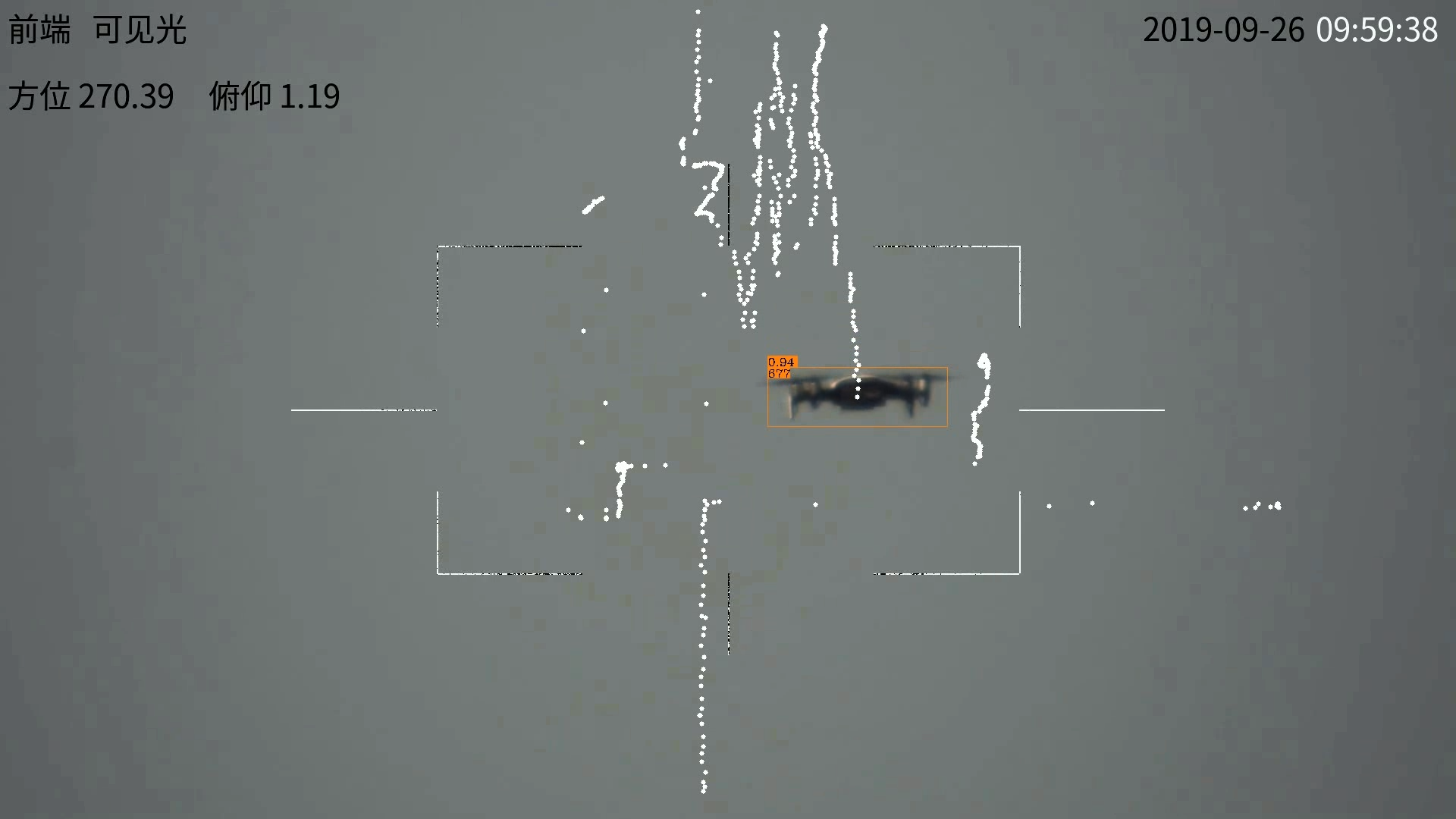}
		\centering
		\caption{Anti-UAV (RGB)}
		\label{Fig:rgbtrack}
	\end{subfigure}
	\begin{subfigure}[b]{0.22\textwidth}
		\includegraphics[width=\linewidth]{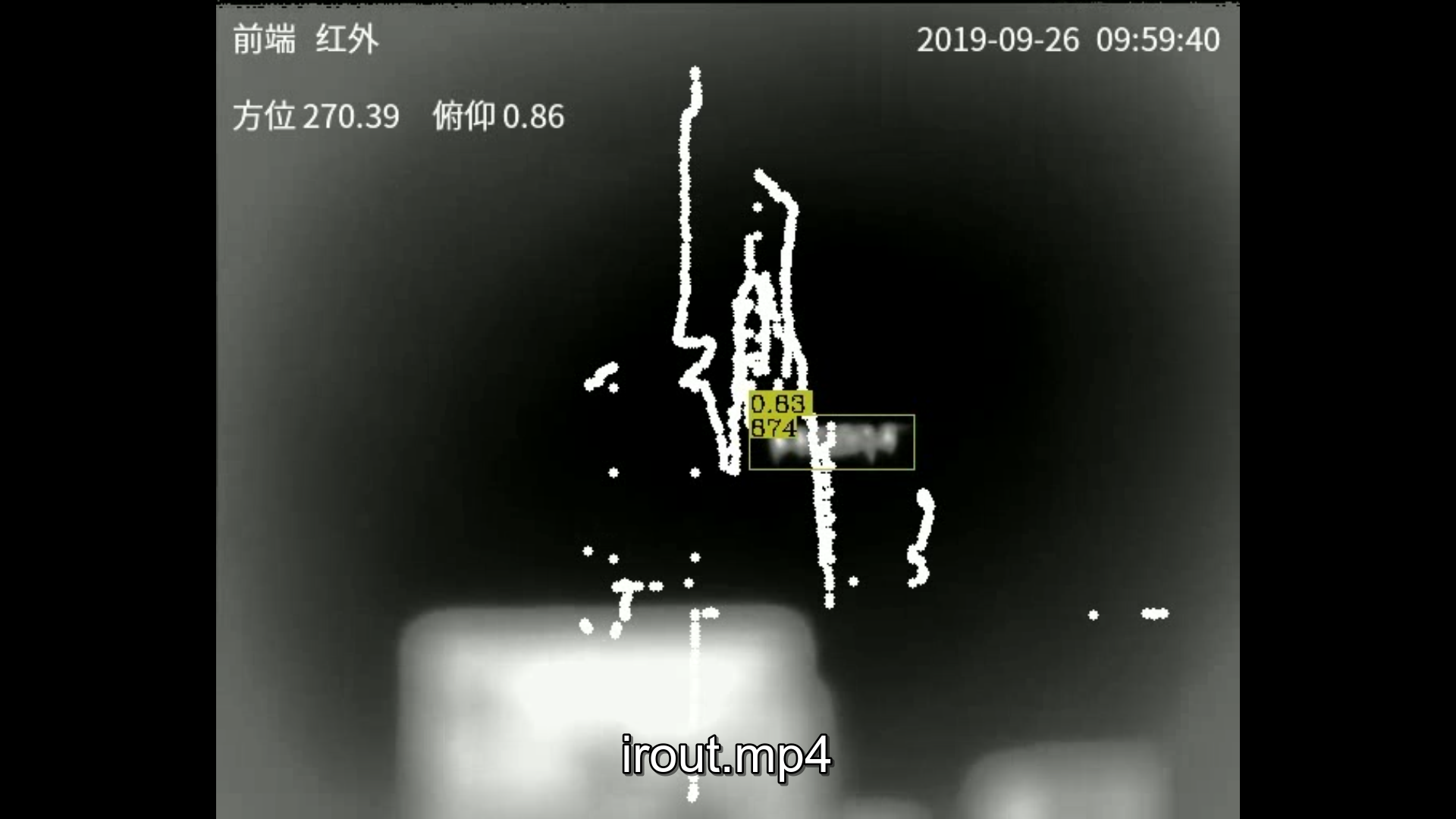}
		\centering
		\caption{Anti-UAV (Infrared)}
		\label{Fig:irtrack}
	\end{subfigure}
	\caption{Examples from each dataset of visualization of drone position history }
	\label{Fig:tracking_example}
\end{figure*}

Interestingly, tracking accuracy is generally similar across all three tracking frameworks, as shown by the MOTA results in Table \ref{Table:tracking_results_deepsort}; differences in accuracy between trackers is less apparent than the difference afforded by different detection architectures. We can attribute this behaviour to two primary causes. Firstly, the Kalman filter within the SORT variants, and the motion model within Tracktor are unable to account for the sudden camera movements within the videos in the dataset. As such, we suspect the differences in trackers are ameliorated when they all fail to reidentify UAV when the camera stops moving. This poses a particular problem for Tracktor, where the regions of interest proposed by the detector and linear velocity model (even after camera motion compensation), are very far from where the UAV will be in the next view. Therefore, the regressor would be unable to locate the new position of the UAV. The challenge can be visualized within Figure \ref{Fig:birdtrack}, where the long, rigid lines white lines indicate where the tracker lost the UAV after the sharp movement of the camera. On the other hand, the light camera shake (the jagged lines within the figure) does not present a problem, through which the UAV is successfully tracked.

Secondly, the reidentification network used within DeepSORT and Tracktor is unable to provide an effective appearance descriptor for UAV, particularly for infrared images. The descriptor is unable to reduce id-switching within tracking (the DeepSORT `ID Sw.' column in table \ref{Table:tracking_results_deepsort} exceeds that of SORT), nor can it contribute to reducing occlusion introduced by other objects within the dynamic environment. As such, the difference in accuracy between tracking frameworks is minimized. In many cases in fact, the re-id network serves to reduce accuracy, where the SORT algorithm resultantly exceeds the DeepSORT algorithm with regards to MOTA. This phenomenon also partially explains how Tracktor yields superior IDF1 performance, which only uses the weak reidentification model to associate new, and previously killed tracks. DeepSORT uses the appearance descriptor at each and every time step however, as part of the distance metric within the Hungarian algorithm. Nevertheless, these results serve to demonstrate the importance of the re-identification network within the tracking pipeline. Investigation of improved re-identification for counter-UAV systems is subsequently identified as an area for future work.

\begin{table*}[!t]
    \centering
    \caption{Cross modality detection performance}
    \resizebox{0.8\linewidth}{!}{%
    \label{Table:rgb_vs_ir_detection}
    \begin{tabular}{c | c | c c c c c c| c c c c}
    \Xhline{2\arrayrulewidth}
    \textbf{Dataset} & \textbf{Model} & $\bf{AP}$ & $\bf{AP_{0.5}}$ & $\bf{AP_{0.75}}$ & $\bf{AP_S}$ & $\bf{AP_{M}}$ & $\bf{AP_{L}}$ 
    & $\bf{AR}$ & $\bf{AR_{S}}$ & $\bf{AR_M}$ & $\bf{AR_L}$ \\
    \Xhline{2\arrayrulewidth}
    \multirow{4}{*}{Anti-UAV RGB} & IR Faster RCNN & \bf 0.394  & \bf 0.828  & \bf 0.322  & 0.000 & \bf 0.372  & \bf 0.481  & \bf 0.484  & 0.000  & \bf 0.453  & \bf 0.598  \\
    
    & IR SSD512 & 0.161  & 0.342  & 0.124 & 0.000  & 0.126  & 0.291  & 0.204  & 0.016  & 0.165  & 0.341  \\
    
    & IR YOLOv3 & 0.113  & 0.299  & 0.061 & 0.002  & \bf 0.082  & 0.240  & 0.156 & \bf 0.019  & 0.119  & 0.286  \\
    
    & IR DETR & 0.284  & 0.728  & 0.148 & 0.000  & 0.241  & 0.409  & 0.390  & 0.000  & 0.354  & 0.519  \\
    \hline
    \multirow{4}{*}{Anti-UAV IR} & RGB Faster RCNN & \bf 0.321  & \bf 0.644  & \bf 0.281  & \bf 0.241 & \bf 0.371  & -  & \bf 0.488  & \bf 0.453  & \bf 0.515  & -  \\
    
    & RGB SSD512 & 0.169  & 0.452  & 0.098 & 0.133  & 0.213  & -  & 0.344  & 0.312  & 0.370  & -  \\
    
    & RGB YOLOv3 & 0.215  & 0.475  & 0.164 & 0.138  & 0.303  & -  & 0.303  & 0.228  & 0.363  & -  \\
    
    & RGB DETR & 0.149  & 0.333  & 0.109 & 0.125  & 0.170  & -  & 0.216  & 0.190  & 0.236  & -  \\
    \Xhline{2\arrayrulewidth}
    \end{tabular}%
		}
\end{table*}

\begin{table*}[!t]
		\centering
		\caption{Object Tracking Performance Benchmark}
		\label{Table:tracking_results_deepsort}
		    \resizebox{0.8\linewidth}{!}{%
		\begin{tabular}{c | c | c c c}
			\Xhline{2\arrayrulewidth}
			\multirow{2}{*}{\textbf{Dataset}} & \multirow{2}{*}{\textbf{Model}} & $\bf{MOTA \uparrow}$ & $\bf{IDF1 \uparrow}$ & $\bf{ID Sw. \downarrow}$\\
			& & \multicolumn{3}{c}{SORT / DeepSORT / Tracktor} \\
			\Xhline{2\arrayrulewidth}
			\multirow{4}{*}{MAV-VID} & Faster RCNN & 0.952 / 0.951 / \textbf{0.955} & 0.547 / 0.591 / 0.680 & 54 / 69 / \textbf{15} \\
			
			& SSD512 & 0.865 / 0.851 / 0.888 & 0.402 / 0.430 / 0.944 & 283 / 435 / 35 \\
			
			& YOLOv3 & 0.881 / 0.877 / 0.934 & 0.260 / 0.419 / 0.970 & 269 / 309 / 26 \\
			
			& DETR & 0.943 / 0.944 / 0.947 & 0.600 / 0.603 / \textbf{0.975} & 61 / 50 / 16 \\
			\hline 
			\multirow{4}{*}{Drone-vs-Bird} & Faster RCNN & 0.420 / 0.415 / 0.459 & 0.377 / 0.391 / 0.439 & 35183 / 35854 / 21952 \\
			
			& SSD512 & 0.364 / 0.328 / \textbf{0.525}  & 0.304 / 0.335 / \textbf{0.872} & 35421 / 39532 / 16517 \\

			& YOLOv3 & 0.400 / 0.370 / 0.523 & 0.287 / 0.359 / 0.868 & 38451 / 41887 / 19623 \\
			
			& DETR & 0.012 / 0.090 / 0.130 & 0.020 / 0.303 / 0.377 & 29126 / 29407 / \textbf{15316} \\
			\hline 
			\multirow{4}{*}{Anti-UAV Full} & Faster RCNN & 0.878 / 0.835 / 0.950 & 0.445 / 0.478 / 0.790 & 2620 / 4094 / 179\\
			
			& SSD512 & 0.856 / 0.806 / 0.936 & 0.373 / 0.406 / 0.966 & 2831 / 4509 / 93\\
			
			& YOLOv3 & 0.767 / 0.693 / \textbf{0.959} & 0.193 / 0.292 / \textbf{0.979} & 3582 / 6097 / 98\\
			
			& DETR & 0.749 / 0.708 / 0.830 & 0.405 / 0.451 / 0.920 & 2814 / 4214 / \textbf{40} \\
			\hline 
			\multirow{4}{*}{Anti-UAV RGB} & Faster RCNN & 0.791 / 0.717 / 0.912 & 0.295 / 0.290 / 0.702 & 2042 / 3163 / 163 \\
			
			& SSD512 & 0.771 / 0.696 / 0.900 & 0.271 / 0.306 / 0.946 & 2096 / 3239 / 126 \\
			
			& YOLOv3 & 0.714 / 0.633 / 0.897 & 0.150 / 0.227 / 0.948 & 2390 / 3617 / 87 \\
			
			& DETR & 0.803 / 0.753 / \textbf{0.940} & 0.323 / 0.327 / \textbf{0.969} & 2101 / 2857 / \textbf{19} \\
			\hline 
			\multirow{4}{*}{Anti-UAV IR} & Faster RCNN & 0.954 / 0.919 / \textbf{0.987} & 0.528 / 0.571 / 0.873 & 560 / 1154 // 22 \\
						
			& SSD512 & 0.922 / 0.904 / 0.963 & 0.442 / 0.454 / 0.981 & 716 / 1020 / 20 \\
			
			& YOLOv3 & 0.780 / 0.718 / 0.985 & 0.180 / 0.327 / \textbf{0.992} & 1216 / 2259 / \textbf{10} \\
			
			& DETR & 0.951 / 0.930 / 0.984  & 0.552 / 0.628 / 0.991 & 575 / 927 / 20\\
			\hline 
		\end{tabular}%
		}
\end{table*}
\section{Conclusion}
\vspace{-2mm}
Computer vision for UAV is an exciting but underdeveloped research field, relative to the necessity of aviation safety and the potential threat (advertent or inadvertent) that UAV carry. To accelerate research in this field, we have conducted a benchmark study on UAV detection and tracking. We have performed an exhaustive evaluation over well-established generic object detectors on three UAV datasets. To the best of our knowledge, we are the first to evaluate a comprehensive array of datasets, across environments, and conditions, captured by both ground and UAV-mounted cameras, optical and IR cameras, at near and far viewpoints. As such, we are able to show that common neural network architectures are capable of detecting small, fast moving objects. Furthermore, YOLOv3 overall yields the best precision (as high as 0.986 mAP), but Faster RCNN is best suited towards detecting small UAV, for which category it consistently achieves the highest mAP (up to 0.770). This is a critical observation when we consider that detecting UAV early is often a stronger priority than detecting precisely. We also show that cross-modality detection is possible for optical and infrared images, wherein training on IR imagery for inference upon visible band spectrum data is not only sufficient but yields excellent performance (0.828 mAP). Furthermore, our tracking results corroborate the suitability towards employing DETR within counter-UAV systems; DETR is best suited for a detection backbone to tracking systems for small objects, and performs adequately upon cross-modal videos. We further demonstrate that the Tracktor framework is best able to track UAV (MOTA: 0.987).

This study serves as a baseline for the community to work towards UAV-specific detectors and trackers. We identify four future research directions to improve counter-UAV systems: 
1) improving long-distance detection ability for when UAV appear small; 2) cross modality training to enforce learning UAV shape 3) bespoke re-identification networks for UAV; 4) improving available datasets. 
Nevertheless, we have demonstrated the efficacy of generic object detectors within both detection and tracking frameworks, upon which bespoke architectures can only improve.

{\small
\bibliographystyle{ieee_fullname}
\bibliography{ref/refs}
}

\end{document}